%% file: emnlp2020.tex
\documentclass[11pt,a4paper]{article}
\usepackage[hyperref]{emnlp2020}
\usepackage{times}
\usepackage{latexsym}

\usepackage{array}
\newcolumntype{P}[1]{>{\centering\arraybackslash}p{#1}}
\newcolumntype{M}[1]{>{\centering\arraybackslash}m{#1}}

\usepackage{microtype}
\usepackage{algorithm} 
\usepackage{algpseudocode}
\usepackage{amsmath}
\usepackage{graphicx}
\usepackage{booktabs,tabularx,enumitem,ragged2e}
\usepackage{comment}
\usepackage{framed}
\usepackage[normalem]{ulem}

\newcommand{\data}{\textsc{ConjNLI}}

\aclfinalcopy 

\title{\textsc{ConjNLI}: Natural Language Inference Over Conjunctive Sentences}

\author{Swarnadeep Saha \;\;\;\;\;\; Yixin Nie \;\;\;\;\;\; Mohit Bansal
\\ 
  UNC Chapel Hill\\ 
  \texttt{\{swarna, yixin1, mbansal\}@cs.unc.edu}
}
  
\date{}

\begin{document}
\maketitle
\begin{abstract}
Reasoning about conjuncts in conjunctive sentences is important for a deeper understanding of conjunctions in English and also how their usages and semantics differ from conjunctive and disjunctive boolean logic. Existing NLI stress tests do not consider non-boolean usages of conjunctions and use templates for testing such model knowledge. Hence, we introduce \data{}, a challenge stress-test for natural language inference over conjunctive sentences, where the premise differs from the hypothesis by conjuncts removed, added, or replaced. These sentences contain single and multiple instances of coordinating conjunctions (``and", ``or", ``but", ``nor") with quantifiers, negations, and requiring diverse boolean and non-boolean inferences over conjuncts. We find that large-scale pre-trained language models like RoBERTa do not understand conjunctive semantics well and resort to shallow heuristics to make inferences over such sentences. As some initial solutions, we first present an iterative adversarial fine-tuning method that uses synthetically created training data based on boolean and non-boolean heuristics. We also propose a direct model advancement by making RoBERTa aware of predicate semantic roles. While we observe some performance gains, \data{} is still challenging for current methods, thus encouraging interesting future work for better understanding of conjunctions.\footnote{\data{} data and code are publicly available at \url{https://github.com/swarnaHub/ConjNLI}.}
\end{abstract}

\section{Introduction}
Coordinating conjunctions are a common syntactic phenomenon in English: 38.8\% of sentences in the Penn Tree Bank have at least one coordinating word between ``and'', ``or'', and ``but''~\cite{marcus1993building}. Conjunctions add complexity to the sentences, thereby making inferences over such sentences more realistic and challenging. A sentence can have many conjunctions, each conjoining two or more conjuncts of varied syntactic categories such as noun phrases, verb phrases, prepositional phrases, clauses, etc. Besides syntax, conjunctions in English have a lot of semantics associated to them and different conjunctions (``and'' vs ``or'') affect the meaning of a sentence differently. 

\begin{table*}[t]
\small
\centering\begin{tabular}{m{0.2cm} | m{5.5cm}| m{5.5cm} | >{\centering\arraybackslash} m{2cm} }
\toprule
\textbf{\#} & \textbf{Premise} & \textbf{Hypothesis} & \textbf{Label} \\ \midrule
\multicolumn{3}{c}{\textbf{\data{} Dataset}} \\ \midrule
1 & He is \textbf{a Worcester resident} and \textbf{a member of the Democratic Party.} & He is a member of the Democratic Party.                          & entailment \\\midrule
2 & He is a member of the Democratic Party.& He is \textbf{a Worcester resident} and \textbf{a member of the Democratic Party}. & neutral \\\midrule
3 & He is \textbf{a Worcester resident} and \textbf{a member of the Democratic Party}.& He is \textbf{a Worcester resident} and \textbf{a member of the Republican Party}. & contradiction   \\\midrule
4 & \textbf{A total of 793880 acre}, or \textbf{36 percent of the park} was affected by the wildfires.&	A total of 793880 acre, was affected by the wildfires. & entailment \\\midrule
5 & Its total running time is \textbf{9 minutes} and \textbf{9 seconds}, spanning seven tracks.& Its total running time is 9 minutes, spanning seven tracks. & contradiction\textsuperscript{$\dagger$} \\\midrule
6 & He began recording for the Columbia Phonograph Company, in \textbf{1889} or \textbf{1890}.& He began recording for the Columbia Phonograph Company, in 1890. & neutral\textsuperscript{$\dagger$} \\\midrule
7 & Fowler wrote or co-wrote \textbf{all} but \textbf{one of} the songs on album.& Fowler wrote or co-wrote all of the songs on album. & contradiction\textsuperscript{$\dagger$} \\\midrule
8 & All devices they tested did not produce \textbf{gravity} or \textbf{anti-gravity}.& All devices they tested did not produce gravity. & entailment \\\midrule
\multicolumn{4}{c}{\textbf{SNLI Dataset}} \\ \midrule
9 & A woman with a green headscarf, blue shirt and a very big grin.& The woman is young. & neutral \\\midrule
\multicolumn{4}{c}{\textbf{MNLI Dataset}} \\ \midrule
10 & You and your friends are not welcome here, said Severn. & Severn said the people were not welcome there. & entailment \\ \bottomrule
\end{tabular}
\vspace{-5pt}
\caption{\label{samples} Examples from our \data{} dataset consisting of single and multiple occurrences of different coordinating conjunctions (and, or, but), boolean or non-boolean in the presence of negations and quantifiers. Typical SNLI and MNLI examples do not require inference over conjuncts. $\dagger = $ Non-boolean usages of different conjunctions.}
\vspace{-7pt}
\end{table*}

Recent years have seen significant progress in the task of Natural Language Inference (NLI) through the development of large-scale datasets like SNLI \cite{bowman2015large} and MNLI \cite{williams2017broad}. Although large-scale pre-trained language models like BERT \cite{devlin2018bert} and RoBERTa \cite{liu2019roberta} have achieved super-human performances on these datasets, there have been concerns raised about these models exploiting idiosyncrasies in the data using tricks like pattern matching \cite{mccoy2019right}. Thus, various stress-testing datasets have been proposed that probe NLI models for simple lexical inferences \cite{glockner2018breaking}, quantifiers \cite{geiger2018stress}, numerical reasoning, antonymy and negation \cite{naik2018stress}. However, despite the heavy usage of conjunctions in English, there is no specific NLI dataset that tests their understanding in detail. Although SNLI has 30\% of samples with conjunctions, most of these examples do not require inferences over the conjuncts that are connected by the coordinating word. On a random sample of 100 conjunctive examples from SNLI, we find that 72\% of them have the conjuncts unchanged between the premise and the hypothesis (e.g., ``Man and woman sitting on the sidewalk'' $\rightarrow$ ``Man and woman are sitting'') and there are almost no examples with non-boolean conjunctions (e.g., ``A total of five men and women are sitting." $\rightarrow$ ``A total of 5 men are sitting." (contradiction)). As discussed below, inference over conjuncts directly translates to boolean and non-boolean semantics and thus becomes essential for understanding conjunctions. 

In our work, we introduce \data{}, a new stress-test for NLI over diverse and challenging conjunctive sentences. Our dataset contains annotated examples where the hypothesis differs from the premise by either a conjunct removed, added or replaced. These sentences contain single and multiple instances of coordinating conjunctions (and, or, but, nor)  with  quantifiers, negations,  and requiring diverse boolean and non-boolean inferences over conjuncts. Table \ref{samples} shows many examples from \data{} and compares these with typical conjunctive examples from SNLI and MNLI. In the first two examples, the conjunct ``a Worcester resident'' is removed and added, while in the third example, the other conjunct ``a member of the Democratic Party'' is replaced by ``a member of the Republican Party''.
Distribution over conjuncts in a conjunctive sentence forms multiple simple sentences. For example, the premise in the first example of Table \ref{samples} can be broken into ``He is a Worcester resident.'' and ``He is a member of the Democratic Party.''. Correspondingly, from boolean semantics, it requires an inference of the form ``$A$ and $B$ $\rightarrow$ $A$''. Likewise, the third example is of the form ``$A$ and $B$ $\rightarrow$ $A$ and $C$''. While such inferences are rather simple from the standpoint of boolean logic, similar rules do not always translate to English, e.g., in non-boolean cases, i.e., an inference of the form ``$A$ and $B$ $\rightarrow$ $A$'' is not always entailment or an inference of the form ``$A$ or $B$ $\rightarrow$ $A$'' is not always neutral \cite{hoeksema1988semantics}. Consider the three examples marked with a $\dagger$ in Table \ref{samples} showing non-boolean usages of ``and'', ``or'' and ``but'' in English. In the fifth example, the total time is a single entity and cannot be separated in an entailed hypothesis.\footnote{While ``9 minutes" can be a vague approximation of ``9 minutes and 9 seconds" in some scenarios \cite{channell1983vague,lasersohn1999pragmatic,kennedy2007vagueness,lee-2008-scalar,morzycki2011metalinguistic}, in our work, we focus on the ``literal interpretations" of sentences with conjunctions.} In the sixth example, ``or" is used as ``exclusive-or" because the person began recording in either 1889 or 1890.

We observe that state-of-the-art models such as BERT and RoBERTa, trained on existing datasets like SNLI and MNLI, often fail to make these inferences for our dataset. For example, BERT predicts entailment for the non-boolean ``and'' example \#5 in Table \ref{samples} as well. This relates to the lexical overlap issue in these models \cite{mccoy2019right}, since all the words in the hypothesis are also part of the premise for the example. Conjunctions are also challenging in the presence of negations. For example, a sentence of the form ``not $A$ or $B$'' translates to ``not $A$ and not $B$'', as shown in example \#8 of Table \ref{samples}. Finally, a sentence may contain multiple conjunctions (with quantifiers), further adding to the complexity of the task (example \#7 in Table \ref{samples}). Thus, our \data{} dataset presents a new and interesting real-world challenge task for the community to work on and allow development of deeper NLI models.

We also present some initial model advancements that attempt to alleviate some of these challenges in our new dataset. First, we create synthetic training data using boolean and non-boolean heuristics. We use this data to adversarially train RoBERTa-style models by an iterative adversarial fine-tuning method. Second, we make RoBERTa aware of predicate semantic roles by augmenting the NLI model with the predicate-aware embeddings of the premise and the hypothesis. Predicate arguments in sentences can help distinguish between two syntactically similar inference pairs with different target labels (Table~\ref{tab:srl} shows an example). Overall, our contributions are:

\begin{itemize}[nosep, wide=0pt, leftmargin=*, after=\strut]
    \item We introduce \data{}, a new stress-test for NLI in conjunctive sentences, consisting of boolean and non-boolean examples with single and multiple coordinating conjunctions (``and'', ``or'', ``but'', ``nor''), negations, quantifiers and requiring diverse inferences over conjuncts (with high inter-annotator agreement between experts).
    \item We show that BERT and RoBERTa do not understand conjunctions well enough and use shallow heuristics for inferences over such sentences.
    \item We propose initial improvements for our task by adversarially fine-tuning RoBERTa using an iterative adversarial fine-tuning algorithm and also augmenting RoBERTa with predicate-aware embeddings. We obtain initial gains but with still large room for improvement, which will hopefully encourage future work on better understanding of conjunctions.
\end{itemize}

\section{Related Work}

\begin{figure*}
    \centering
    \includegraphics[width=\textwidth]{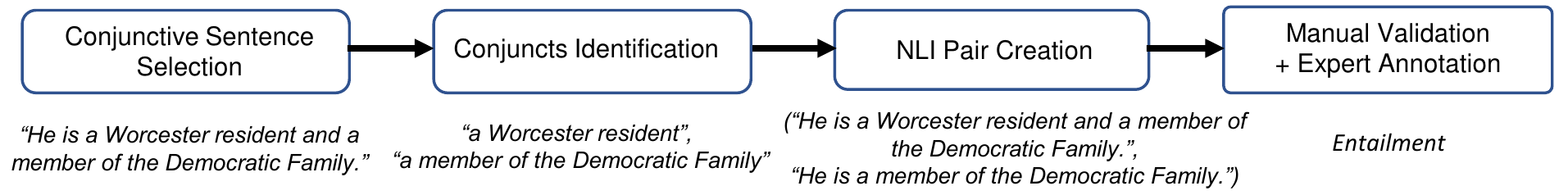}
    \caption{\centering Flow diagram of \data{} dataset creation.}
    \label{fig:data_flow}
    \vspace{-7pt}
\end{figure*}{}

Our work is positioned at the intersection of understanding the semantics of conjunctions in English and its association to NLI. 

\paragraph{Conjunctions in English.}

There is a long history of analyzing the nuances of coordinating conjunctions in English and how these compare to boolean and non-boolean semantics \cite{gleitman1965coordinating,keenan2012boolean,schein2017and}. Linguistic studies have shown that noun phrase conjuncts in ``and'' do not always behave in a boolean manner \cite{massey1976tom, hoeksema1988semantics, krifka1990boolean}. In the NLP community, studies on conjunctions have mostly been limited to treating it as a syntactic phenomenon. One of the popular tasks is that of conjunct boundary identification \cite{agarwal1992simple}. \citet{ficler2016coordination} show that state-of-the-art parsers often make mistakes in identifying conjuncts correctly and develop neural models to accomplish this \cite{ficler2016neural, teranishi2019decomposed}. \citet{saha2018open} also identify conjuncts to break conjunctive sentences into simple ones for better downstream Open IE \cite{banko2007open}. \citet{aixiu2019representation} study French and English coordinated noun phrases and investigate whether neural language models can represent constituent-level features and use them to drive downstream expectations. In this work, we broadly study the semantics of conjunctions through our challenging dataset for NLI.

\paragraph{Analyzing NLI Models.}

Our research follows a body of work trying to understand the weaknesses of neural models in NLI. \citet{poliak2018hypothesis, gururangan2018annotation} first point out that hypothesis-only models also achieve high accuracy in NLI, thereby revealing weaknesses in existing datasets. Various stress-testing and analysis datasets have been proposed since, focusing on lexical inferences \cite{glockner2018breaking}, quantifiers \cite{geiger2018stress}, biases on specific words \cite{carmona2018behavior}, verb verdicality \cite{ross2019well}, numerical reasoning \cite{ravichander2019equate}, negation, antonymy \cite{naik2018stress}, pragmatic inference \cite{jeretic2020natural}, systematicity of monotonicity \cite{yanaka2020neural}, and linguistic minimal pairs \cite{warstadt2020blimp}. Besides syntax, other linguistic information have also been investigated  \cite{poliak2018collecting, white2017inference} but none of these focus on conjunctions. The closest work on conjunctions is by \citet{richardson2019probing} where they probe NLI models through semantic fragments. However, their focus is only on boolean ``and'', allowing them to assign labels automatically through simple templates. Also, their goal is to get BERT to master semantic fragments, which, as they mention, is achieved with a few minutes of additional fine-tuning on their templated data. \data{}, however, is more diverse and challenging for BERT-style models, includes all common coordinating conjunctions, and captures non-boolean usages.

\paragraph{Adversarial Methods in NLP.} Adversarial training for robustifying neural models has been proposed in many NLP tasks, most notably in QA \cite{jia2017adversarial, wang2018robust} and NLI \cite{nie2019analyzing}. \citet{nie2019adversarial} improve existing NLI stress tests using adversarially collected NLI data (ANLI) and \citet{Kaushik2020Learning} use counter-factually augmented data for making models robust to spurious patterns. Following \citet{jia2017adversarial}, we also create adversarial training data by performing all data creation steps except for the expensive human annotation. Our iterative adversarial fine-tuning method adapts adversarial training in a fine-tuning setup for BERT-style models and improves results on \data{} while maintaining performance on existing datasets.

\section{Data Creation}
\label{data-creation}

Creation of \data{} involves four stages, as shown in Figure \ref{fig:data_flow}. The (premise, hypothesis) pairs are created automatically, followed by manual verification and expert annotation.

\subsection{Conjunctive Sentence Selection}
We start by choosing conjunctive sentences from Wikipedia containing all common coordinating conjunctions (``and'', ``or'', ``but'', ``nor''). Figure \ref{fig:data_flow} shows an example. We choose Wikipedia because it contains complex sentences with single and multiple conjunctions, and similar choices have also been made in prior work on information extraction from conjunctive sentences \cite{saha2018open}. In order to capture a diverse set of conjunctive phenomena, we gather sentences with multiple conjunctions, negations, quantifiers and various syntactic constructs of conjuncts.

\subsection{Conjuncts Identification}
For conjunct identification, we process the conjunctive sentence using a state-of-the-art constituency parser implemented in AllenNLP\footnote{\em https://demo.allennlp.org/constituency-parsing} and then choose the two phrases in the resulting constituency parse on either side of the conjunction as conjuncts. A conjunction can conjoin more than two conjuncts, in which case we identify the two surrounding the conjunction and ignore the rest. Figure \ref{fig:data_flow} shows an example where the two conjuncts ``a Worcester resident'' and ``a member of the Democratic Party'' are identified with the conjunction ``and''.

\subsection{NLI Pair Creation}
Once the conjuncts are identified, we perform three operations by removing, adding or replacing one of the two conjuncts to obtain another sentence such that the original sentence and the modified sentence form a plausible NLI pair. Figure \ref{fig:data_flow} shows a pair created by the removal of one conjunct. We create the effect of adding a conjunct by swapping the premise and hypothesis from the previous example. We replace a conjunct by finding a conjunct word that can be replaced by its antonym or co-hyponym. Wikipedia sentences frequently contain numbers or names of persons in the conjuncts which are replaced by adding one to the number and randomly sampling any other name from the dataset respectively. We apply the three conjunct operations on all collected conjunctive sentences.

\subsection{Manual Validation \& Expert Annotation}

Since incorrect conjunct identification can lead to the generation of a grammatically incorrect sentence, the pairs are first manually verified for grammaticality. The grammatical ones are next annotated by two  English-speaking experts (with prior experience in NLI and NLP) into entailment, neutral and contradiction labels. We refrain from using Amazon Mechanical Turk for the label assignment because our NLI pairs' labeling requires deeper understanding and identification of the challenging conjunctive boolean versus non-boolean semantics (see examples \#1 and \#5 in Table \ref{samples} where the same conjunct removal operation leads to two different labels). Expert annotation has been performed in previous NLI stress-tests as well \cite{ravichander2019equate, mccoy2019right} so as to ensure a high-quality dataset.

\noindent \textbf{Annotator Instructions and Agreement}: Each annotator is initially trained by showing 10 examples, of which some have boolean usages and others non-boolean. The examples are further accompanied with clear explanations for the choice of labels. The appendix contains a subset of these examples. The annotations are done in two rounds -- in the first, each annotator annotated the examples independently and in the second, the disagreements are discussed to resolve final labels. The inter-annotator agreement between the annotators has a high Cohen's Kappa ($\kappa$) of 0.83  and we keep only those pairs that both agree on.

\section{Data Analysis}

Post-annotation, we arrive at a consolidated set of 1623 examples, which is a reasonably large size compared to previous NLI stress-tests with expert annotations.  We randomly split these into 623 validation and 1000 test examples, as shown in Table \ref{tab:split}. \data{} also replicates the approximate distribution of each conjunction in English (Table \ref{tab:sem_dist}). Thus, ``and'' is maximally represented in our dataset, followed by ``or''\footnote{We consider sentences with ``nor'' as part of ``or''.} and ``but''. Sentences with multiple conjunctions make up a sizeable 23\% of \data{} to reflect real-world challenging scenarios. As we discussed earlier, conjunctions are further challenging in the presence of quantifiers and negations, due to their association with boolean logic. These contribute to 18\% and 10\% of the dataset, resp.

\begin{table}[t]
\centering
\small
\begin{tabular}{lrrrr}
\toprule
              & Ent & Neu & Contra & Total\\ \midrule
Conj Dev   & 204 &	281 &	138 &	623                     \\ 
Conj Test       & 332 &	467 &	201 &	1000                    \\ 
Conj All & 536 &	748 &	339 &	1623                    \\ \bottomrule

\end{tabular}
\caption{\label{tab:split} Dataset splits of \data{}.}
\end{table}

\begin{table}[t]
\centering
\small
\begin{tabular}{lrrrrrr}
\toprule
              & and & or & but & multiple & quant & neg \\ \midrule
Conj Dev    & 320 &	293 &	99 & 152 & 131 & 70                    \\ 
Conj Test       & 537 &	471 &	135 & 229 & 175 & 101                   \\ 
Conj All & 857 &	764 &	234 & 381 & 306 & 171                  \\ \bottomrule                  
\end{tabular}
\vspace{-5pt}
\caption{\label{tab:sem_dist} Data analysis by conjunction types, presence of quantifiers and negations.}
\end{table}

\begin{table}[t]
\small
\centering\begin{tabular}{m{5cm}| >{\centering\arraybackslash} m{2cm} }
\toprule
\textbf{Sentence} & \textbf{CT} \\ \midrule
Historically, the Commission was run by \textbf{three commissioners} or \textbf{fewer}.                          & NP + Adj \\\midrule
\textbf{Terry Phelps} and \textbf{Raffaella Reggi} were the defending champions but did not compete that year. & NP + NP \\\midrule
Terry Phelps and Raffaella Reggi \textbf{were the defending champions} but \textbf{did not compete that year}. & VP + VP   \\\midrule
It is for Orienteers \textbf{in} or \textbf{around} North Staffordshire and South Cheshire. & Prep + Prep \\\midrule
\textbf{It is a white solid}, but \textbf{impure samples can appear yellowish}. & Clause + Clause \\\midrule
Pantun were originally not written down, the bards often being \textbf{illiterate} and \textbf{in many cases blind}. & Adj + PP\\\midrule
A queue is \textbf{an example of a linear data structure}, or \textbf{more abstractly a sequential collection}. & NP + AdvP \\\bottomrule
\end{tabular}
\vspace{-5pt}
\caption{\label{tab:syn} \data{} sentences consist of varied syntactic conjunct categories (bolded). CT = Conjunct Types, NP = Noun Phrase, VP = Verb Phrase, AdvP = Adverbial Phrase.}
\vspace{-5pt}
\end{table}

We note that conjunctive sentences can contain conjuncts of different syntactic categories, ranging from words of different part of speech tags to various phrasal constructs to even sentences. Table \ref{tab:syn} shows a small subset of the diverse syntactic constructs of conjuncts in \data{}. The conjuncts within a sentence may belong to different categories -- the first example conjoins a noun phrase with an adjective. Each conjunct can be a clause, as shown in the fifth example.

\subsection{Iterative Adversarial Fine-Tuning}
\label{adv-data}
\paragraph{Automated Adversarial Training Data Creation.} Creation of large-scale conjunctive-NLI training data, where each example is manually labeled, is prohibitive because of the amount of human effort involved in the process and the diverse types of exceptions involved in the conjunction inference labeling process. Hence, in this section, we first try to automatically create some training data to train models for our challenging \data{} stress-test and show the limits of such rule-based adversarial training methods. For this automated training data creation, we follow the same process as Section~\ref{data-creation} but replace the expert human-annotation phase with automated boolean rules and some initial heuristics for non-boolean\footnote{Non-boolean usages of conjunctions, to the best of our knowledge, cannot be identified automatically; and in fact, that is the exact motivation of \textsc{ConjNLI}, which encourages the development of such models.} semantics so as to assign labels to these pairs automatically. For ``boolean and'', if ``A and B'' is true, we assume that A and B are individually true, and hence whenever we remove a conjunct, we assign the label entailment and whenever we add a conjunct, we assign the label neutral. Examples with conjunct replaced are assigned the label contradiction. As already shown in Table \ref{samples}, there are of course exceptions to these rules, typically arising from the ``non-boolean'' usages. \citet{hoeksema1988semantics, krifka1990boolean} show that conjunctions of proper names or named entities, definite descriptions, and existential quantifiers often do not behave according to general boolean principles. Hence, we use these suggestions to develop some initial non-boolean heuristics for our automated training data creation. First, whenever we remove a conjunct from a named entity (``Franklin and Marshall College'' $\rightarrow$ ``Franklin College''), we assign the label neutral because it typically refers to a different named entity. Second, ``non-boolean and'' is prevalent in sentences where the conjunct entities together map onto a collective entity and often in the presence of certain trigger words like ``total'', ``group'', ``combined'', etc. (but note that this is not always true). For example, removing the conjunct ``flooding'' in the sentence ``In total, the flooding and landslides killed 3,185 people in China.'' should lead to contradiction. We look for such trigger words in the sentence and heuristically assign contradiction label to the pair.
Like ``and'', the usage of ``or'' in English often differs from boolean ``or''. The appendix contains details of the various interpretations of English ``or'', and our adversarial data creation heuristics.

\begin{algorithm}[t]
\caption{Iterative Adversarial Fine-Tuning}\label{alg:conj-nli}
\begin{algorithmic}[1]
\small
\State $model$ = finetune($\mathit{RoBERTa}$, $\mathit{MNLI_{train}}$)
\State $adv\_train$ = get\_adv\_data()
\State $k$ = len($adv_{train}$)
\For{$e = 1$ to $num\_epochs$}
\State $\mathit{MNLI\_small}$ = sample\_data($\mathit{MNLI_{train}}$, $k$)
\State $all\_data$ = $\mathit{MNLI\_small}$ $\bigcup$ $adv\_train$
\State Shuffle $all\_data$
\State $model$ = finetune($model$, $all\_data$)
\EndFor
\end{algorithmic}
\end{algorithm}

\section{Methods}
In this section, we first describe our iterative adversarial fine-tuning method (including the creation of adversarial training data), followed by some initial predicate-aware models to try to tackle \data{}.

We create a total of 15k adversarial training examples using the aforementioned shallow heuristics, with an equal number of examples for ``and'', ``or'' and ``but''. A random sample of 100 examples consisting of an equal number of ``and'', ``or'' and ``but'' examples are chosen for manual validation by one of the annotators, yielding an accuracy of 70\%. 
We find that most of the errors either have challenging non-boolean scenarios which cannot be handled by our heuristics or have ungrammatical hypotheses, originating from parsing errors.

\begin{figure}[t]
    \centering
    \includegraphics[width=0.48\textwidth]{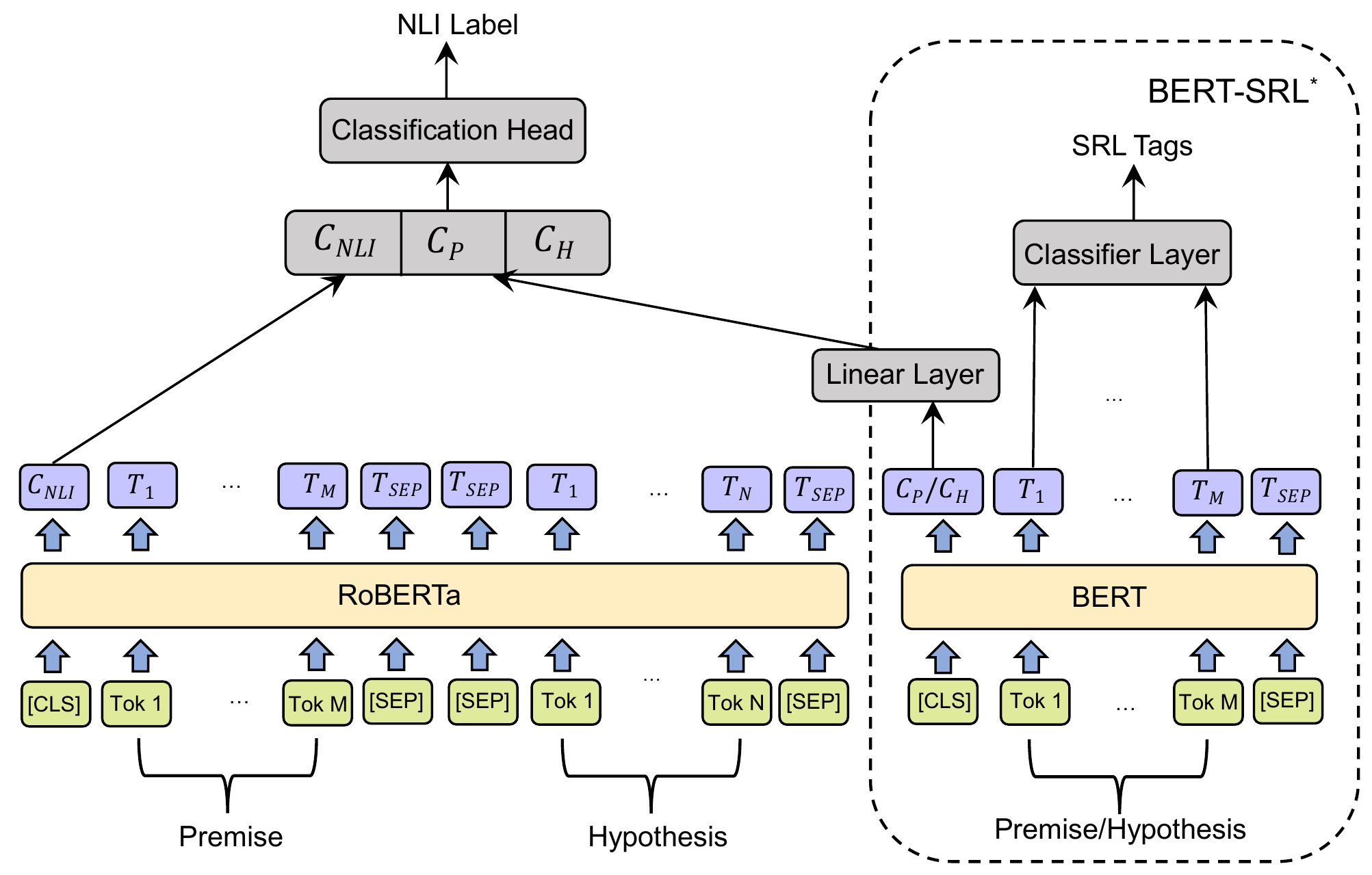}
    \caption{Architecture diagram of predicate-aware RoBERTa model for \data{}. * = BERT-SRL weights are frozen while fine-tuning on the NLI task.}
    \label{fig:model}
\end{figure}{}

\begin{table*}[t]
    \centering
    \small
    \begin{tabular}{|m{4.5cm}|m{4.5cm}|m{1cm}|m{4cm}|}
    \toprule
         \textbf{Premise} & \textbf{Hypothesis} & \textbf{Label} & \textbf{SRL Tags} \\ \midrule
        It premiered on 27 June 2016 and airs Mon-Fri 10-11pm IST. & It premiered on 28 June 2016 and airs Mon-Fri 10-11pm IST. & contra & ARG1:``It'', Verb:``premiered'', Temporal:``on 27 June 2016'' \\ \midrule
        He also played in the East-West Shrine Game and was named MVP of the Senior Bowl. & He also played in the North-South Shrine Game and was named MVP of the Senior Bowl. & neutral & ARG1: ``He'', Discource:``also'', Verb:``played'', Location:``in the East-West Shrine Game''.\\ \bottomrule
    \end{tabular}
    \vspace{-5pt}
    \caption{Two examples from \data{} where SRL tags can help the model predict the correct label.
    \vspace{-5pt}
    }
    \label{tab:srl}
    
\end{table*}{}

\paragraph{Algorithm for Iterative Adversarial Fine-Tuning.}

Our adversarial training method is outlined in Algorithm \ref{alg:conj-nli}. We look to improve results on \data{} through adversarial training while maintaining state-of-the-art results on existing datasets like MNLI. Thus, we first fine-tune RoBERTa on the entire MNLI training data. Next, at each epoch, we randomly sample an equal amount of original MNLI training examples with conjunctions as the amount of adversarial training data. We use the combined data to further fine-tune the MNLI fine-tuned model. At each epoch, we iterate over the original MNLI training examples by choosing a different random set every time, while keeping the adversarial data constant. The results section discusses the efficacy of our algorithm.

As shown later, adversarial training leads to limited improvements on \data{} due to the rule-based training data creation. Since real-world conjunctions are much more diverse and tricky, our dataset encourages future work by the community and also motivates a need for direct model development like our initial predicate-aware RoBERTa.

\subsection{Initial Predicate-Aware (SRL) RoBERTa}
We find that \data{} contains examples where the inference label depends on the predicate and the predicate roles in the sentence. Consider the two examples in Table \ref{tab:srl}. The two premises are syntactically similar and both undergo the conjunct replacement operation for creating the hypothesis. However, their respective predicates ``premiered'' and ``played'' have different arguments, notably one referring to a premier date while the other describing playing in a location. Motivated by the need to better understand predicates and predicate roles in NLI pairs, we propose a predicate-aware RoBERTa model, built on top of a standard RoBERTa model for NLI. Figure \ref{fig:model} shows the architecture diagram. We make the model aware of predicate roles by using representations of both the premise and the hypothesis from a fine-tuned BERT model on the task of Semantic Role Labeling (SRL).\footnote{Our initial experiments show that BERT marginally outperforms RoBERTa for SRL.} Details of the BERT-SRL model can be found in the appendix.
\noindent Let the RoBERTa embedding of the $[CLS]$ token be denoted by $C_{NLI}$. The premise and hypothesis are also passed through the BERT-SRL model to obtain predicate-aware representations for each. These are similarly represented by the corresponding $[CLS]$ token embeddings. We learn a linear transformation on top of these embeddings to obtain $C_P$ and $C_H$. Following \citet{pang2019improving}, where they use late fusion of syntactic information for NLI, we perform the same with the predicate-aware SRL representations. A final classification head gives the predictions.

\begin{table}[t]
\centering
\small
\begin{tabular}{lrrrr}
\toprule
& \multicolumn{1}{c}{MD} & \multicolumn{1}{c}{SD} & \multicolumn{1}{c}{CD}  & \multicolumn{1}{c}{CT}\\ \midrule 

BERT-S         & - & 90.85 & 60.03 & 59.40 \\ 
BERT-M           &	84.10/83.90 & - & 58.10 & 61.40\\ 
RoBERTa-S        & - & \textbf{91.87} & 60.99 & 63.50\\ 
RoBERTa-M       &	\textbf{87.56/87.51} & - &  \textbf{64.68} &	\textbf{65.50}\\ \bottomrule
\end{tabular}
\caption{\label{baseline} Comparison of BERT and RoBERTa trained on SNLI and MNLI and tested on respective dev sets and \data{}. MNLI Dev (MD) results are in match/mismatched format. SD = SNLI Dev, CD = Conj Dev, CT = Conj Test.}
\vspace{-10pt}
\end{table}

\subsection{Predicate-Aware RoBERTa with Adversarial Fine-Tuning}
In the last two subsections, we proposed enhancements both on the data side and the model side to tackle \data{}. Our final joint model now combines predicate-aware RoBERTa with iterative adversarial fine-tuning. We conduct experiments to analyze the effect of each of these enhancements as well as their combination. 

\section{Experiments and Results}
We perform experiments on three datasets - (1) \data{}, (2) SNLI \cite{bowman2015large} and (3) MNLI \cite{williams2017broad}. The appendix contains details about our experimental setup.

\subsection{Baselines}

We first train BERT and RoBERTa on the SNLI (BERT-S, RoBERTa-S) and MNLI (BERT-M, RoBERTa-M) training sets and evaluate their performance on the respective dev sets and \data{}, as shown in Table \ref{baseline}. We observe a similar trend for both MNLI and \data{}, with MNLI-trained RoBERTa being the best performing model. This is perhaps unsurprising as MNLI contains more complex inference examples compared to SNLI. The results on \data{} are however significantly worse than MNLI, suggesting a need for better understanding of conjunctions. We also experimented with older models like ESIM \cite{chen2017enhanced} and the accuracy on \data{} was much worse at 53.10\%. Moreover, in order to verify that our dataset does not suffer from the `hypothesis-only' bias~\cite{poliak2018hypothesis, gururangan2018annotation}, we train a hypothesis-only RoBERTa model and find it achieves a low accuracy of 35.15\%. All our successive experiments are conducted using RoBERTa with MNLI as the base training data, owing to its superior performance.

In order to gain a deeper understanding of these models' poor performance, we randomly choose 100 examples with ``and” and only replace the ``and" with ``either-or" (exclusive-or) along with the appropriate change in label. For example, ``He received bachelor’s degree in 1967 and PhD in 1973." $\rightarrow$ ``He received bachelor’s degree in 1967." (entailment) is changed to ``He either received bachelor’s degree in 1967 or PhD in 1973." $\rightarrow$ ``He received bachelor’s degree in 1967." (neutral). We find that while RoBERTa gets most of the ``and" examples correct, the ``or" examples are mostly incorrect because the change in conjunction does not lead to a change in the predicted label for any of the examples. This points to the lexical overlap heuristic \cite{mccoy2019right} learned by the model that if the hypothesis is a subset of the premise, the label is mostly entailment, while ignoring the type of conjunction.

\begin{table}
\small
\centering
\begin{tabular}{lrrr}
\toprule
                   & \multicolumn{1}{c}{Conj Dev} & \multicolumn{1}{c}{MNLI Dev} & \multicolumn{1}{c}{Conj Test} \\  \midrule
BERT & 58.10 &	84.10/83.90 &	61.40 \\ 
RoBERTa         & 64.68 &	\textbf{87.56/87.51} &	65.50 \\ 
AFT  &	67.57 &	76.61/76.68 &	66.40\\ 
IAFT & \textbf{69.18} &	86.93/86.81 &	\textbf{67.90} \\ \bottomrule
\end{tabular}
\vspace{-5pt}
\caption{\label{adv-comparison} Table showing the effectiveness of IAFT over AFT and other baseline models.}
\vspace{-7pt}
\end{table}

\subsection{Iterative Adversarial Fine-Tuning}

We compare our proposed Iterative Adversarial Fine-Tuning (IAFT) approach with simple Adversarial Fine-Tuning (AFT) wherein we start with the MNLI fine-tuned RoBERTa model and fine-tune it further with the adversarial data, in a two-step process.\footnote{AFT, in principle, is similar to the \texttt{Inoculation by Fine-Tuning} strategy \cite{liu2019inoculation}, with the exception that they inject some examples from the challenge set for training, while we have a separate heuristically-created adversarial training set.} Table \ref{adv-comparison} shows that IAFT obtains the best average results between \data{} and MNLI with 2\% improvement on the former and retaining state-of-the-art results on the latter. In the simple AFT setup, the model gets biased towards the adversarial data, resulting in a significant drop in the original MNLI results. The slight drop in \data{} results also indicates that the MNLI training data is useful for the model to learn about basic paraphrasing skills required in NLI. IAFT achieves that, by mixing the adversarial training data with an equal amount of MNLI examples in every epoch. We also analyze a subset of examples fixed by IAFT and find that unsurprisingly (based on the heuristics used to automatically create the adversarial training data in Sec.~\ref{adv-data}), it corrects more boolean examples than non-boolean (65\% vs 35\%) and the non-boolean examples either have named entities or collective conjuncts.

We note that IAFT is a generic approach and can be used to improve other stress-tests in an adversarial fine-tuning setup. As an example, we apply it on the boolean subset of the dataset by \citet{richardson2019probing} containing samples with ``boolean and" and find that our model achieves a near perfect accuracy on their test set. Specifically, RoBERTa, trained on only MNLI, achieves a low accuracy of 41.5\% on the test set, but on applying IAFT with an equal mix of MNLI and their training data in every epoch, the test accuracy improves to 99.8\%, while also retaining MNLI matched/mismatched results at 86.45/86.46\%.

\begin{table}
\small
\centering
\begin{tabular}{lrrr}
\toprule
                   & \multicolumn{1}{c}{Conj Dev} & \multicolumn{1}{c}{MNLI Dev} & \multicolumn{1}{c}{Conj Test} \\  \midrule
BERT & 58.10 &	84.10/83.90 &	61.40 \\ 
RoBERTa         & 64.68 &	87.56/87.51 &	65.50 \\ 
PA & 64.88 & \textbf{87.75/87.63} &	66.30 \\ 
IAFT & \textbf{69.18} &	86.93/86.81 &	\textbf{67.90} \\ 
PA-IAFT & 68.89 & 87.07/86.93 &	67.10 \\ \bottomrule
\end{tabular}
\vspace{-5pt}
\caption{\label{final-table} Comparison of all our final models on \data{} and MNLI.
}
\vspace{-10pt}
\end{table}

\subsection{Predicate-Aware RoBERTa with Adversarial Fine-Tuning}

Table \ref{final-table} consolidates our final results on both the datasets. We compare the baselines BERT and RoBERTa with (1) Predicate-aware RoBERTa (PA), (2) RoBERTa with Iterative Adversarial Fine-Tuning (IAFT), and (3) Predicate-aware RoBERTa with Iterative Adversarial Fine-tuning (PA-IAFT). Our first observation is that PA marginally improves results on both the datasets. This is encouraging, as it shows that NLI in general, can benefit from more semantic information. However, we obtain a larger gain on \data{} with adversarial training. This, however, is unsurprising as the adversarial training data is specifically curated for the task, whereas PA is only exposed to the original MNLI training data. On combining both, our results do not improve further, thus promoting future work by the community on better understanding of conjunctions. Finally, all our models encouragingly maintain state-of-the-art results on MNLI.

\subsection{Amount of Adversarial Training Data}
\begin{figure}[t]
    \centering
    \includegraphics[width=0.9\columnwidth]{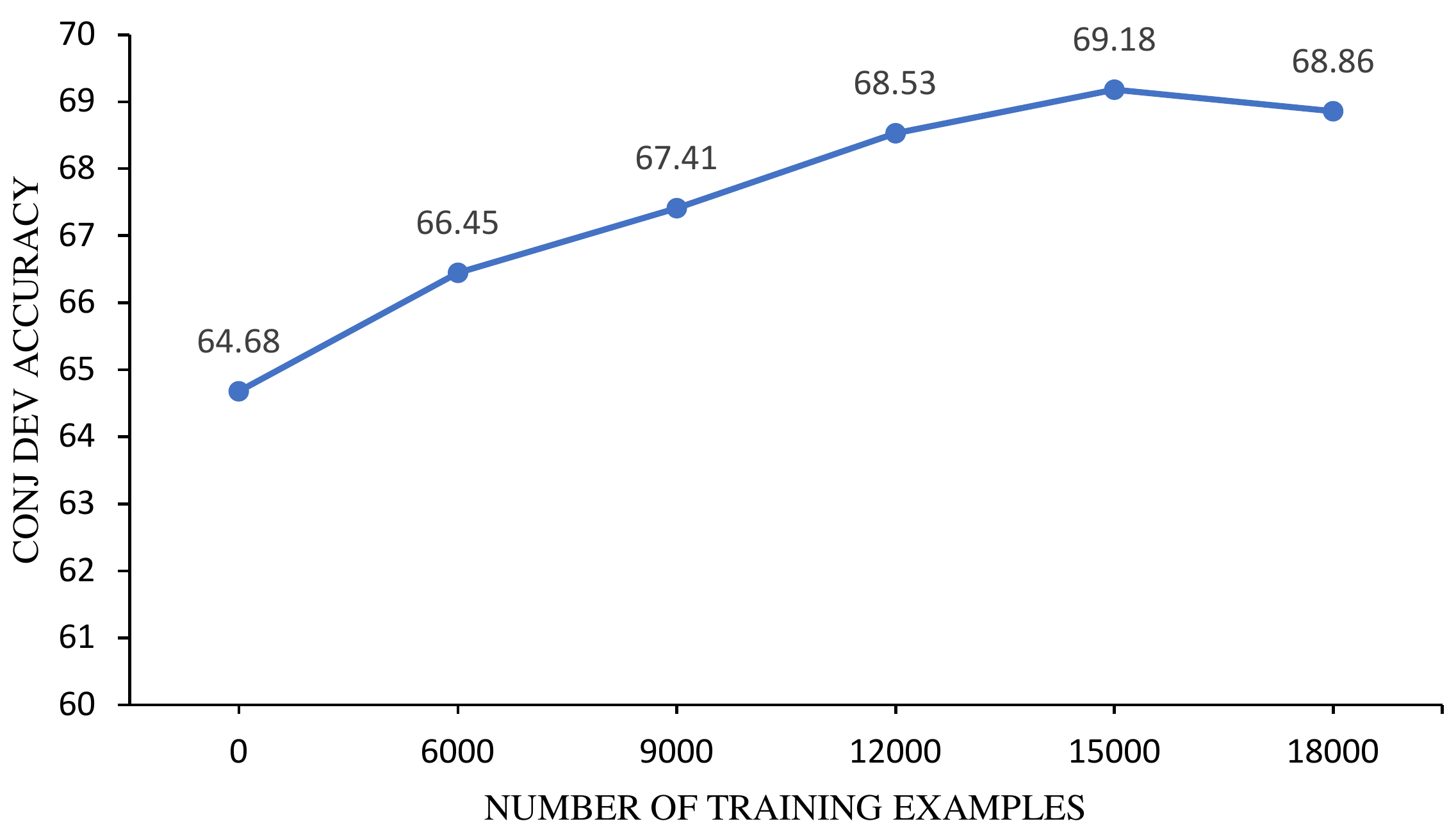}
    \caption{Effect of amount of adversarial training data.}
    \label{fig:data-size}
    \vspace{-7pt}
\end{figure}

We investigate the amount of training data needed for RoBERTa-style models to learn the heuristics used to create the adversarial data. We experiment with the IAFT model on \data{} dev and linearly increase the data size from 6k to 18k, comprising of an equal amount of ``and'', ``or'' and ``but'' examples. Figure \ref{fig:data-size} shows the accuracy curve. We obtain maximum improvements with the first 12k examples (4 points), marginal improvement with the next 3k and a slight drop in performance with the next 3k. Early saturation shows that RoBERTa learns the rules using a small number of examples only and also exposes the hardness of \data{}.

\subsection{Instability Analysis}

\citet{zhou2020curse} perform an in-depth analysis of the various NLI stress tests like HANS \cite{mccoy2019right}, BREAK-NLI \cite{glockner2018breaking}, etc and find that different random initialization seeds can lead to significantly different results on these datasets. They show that this instability largely arises from high inter-example similarity, as these datasets typically focus on a particular linguistic phenomenon by leveraging only a handful of patterns. Thus, following their suggestion, we conduct an instability analysis of \data{} by training RoBERTa on MNLI with 10 different seeds (1 to 10) and find that the results on \data{} are quite robust to such variations. The mean accuracy on \data{} dev is 64.48, with a total standard deviation of 0.59, independent standard deviation of 0.49 and a small inter-data covariance of 0.22. \data{}'s stable results compared to most previous stress-tests indicate the diverse nature of conjunctive inferences captured in the dataset.

\subsection{Analysis by Conjunction Type}
In Table \ref{tab:conj_type}, we analyze the performance of the models on the subset of examples containing ``and'', ``or'', ``but'' and multiple conjunctions. We find that ``or'' is the most challenging for pre-trained language models, particularly because of its multiple interpretations in English. We also note that all models perform significantly better on sentences with ``but'', owing to fewer non-boolean usages in such sentences. Our initial predicate-aware model encouragingly obtains small improvements on all conjunction types (except ``but''), indicating that perhaps these models can benefit from more linguistic knowledge. Although single conjunction examples benefit from adversarial training, multiple conjunctions prove to be challenging mainly due to the difficulty in automatically parsing and creating perfect training examples with such sentences \cite{ficler2016neural}.

\begin{table}
\small
\centering
\begin{tabular}{lrrrrr}
\toprule
                   & \multicolumn{1}{c}{And} & \multicolumn{1}{c}{Or} & \multicolumn{1}{c}{But} & \multicolumn{1}{c}{Multiple} & \multicolumn{1}{c}{All} \\ \midrule 
RoBERTa         & 65.36 &	59.87 &	\textbf{81.48} &	65.93 &	65.60 \\
PA  &	66.29 &	60.93 &	\textbf{81.48} &	\textbf{66.81} &	66.30\\ 
IAFT & \textbf{67.59} &	\textbf{62.20} &	80.00 &	62.88 &	\textbf{67.90} \\ \bottomrule
\end{tabular}
\vspace{-5pt}
\caption{\label{tab:conj_type} Comparison of all models on the subset of each conjunction type of \data{}.}
\vspace{-10pt}
\end{table}

\subsection{Analysis of Boolean versus Non-Boolean Conjunctions}
One of the expert annotators manually annotated the \data{} dev set for boolean and non-boolean examples. We find that non-boolean examples contribute to roughly 34\% of the dataset. Unsurprisingly, all models perform significantly better on the boolean subset compared to the non-boolean one. Specifically, the accuracies for RoBERTa, IAFT and PA on the boolean subset are 68\%, 72\% and 69\% respectively, while on the non-boolean subset, these are 58\%, 61\% and 58\% respectively. Based on these results, we make some key observations: (1) Non-boolean accuracy for all models are about 10\% less than the boolean counterpart, revealing the hardness of the dataset, (2) IAFT improves both boolean and non-boolean subsets because of the non-boolean heuristics used in creating its adversarial training data, (3) PA only marginally improves the boolean subset, suggesting the need for better semantic models in future work. In fact, \data{} also provides a test bed for designing good semantic parsers that can automatically distinguish between boolean and non-boolean conjunctions.

\section{Conclusion}
We presented \data{}, a new stress-test dataset for NLI in conjunctive sentences (``and'', ``or'', ``but'', ``nor'') in the presence of negations and quantifiers and requiring diverse ``boolean'' and ``non-boolean" inferences over conjuncts. Large-scale pre-trained LMs like RoBERTa are not able to optimally understand the conjunctive semantics in our dataset. We presented some initial solutions via adversarial training and a predicate-aware RoBERTa model, and achieved some reasonable performance gains on \data{}. However, we also show limitations of our proposed methods, thereby encouraging future work on \data{} for better understanding of conjunctive semantics.

\section*{Acknowledgements}

We thank the reviewers and Adina Williams, Peter Hase, Yichen Jiang, and Prof. Katya Pertsova (UNC Linguistics department) for their helpful suggestions, and the experts for data annotation. This work was supported by DARPA MCS Grant N66001-19-2-4031, NSF-CAREER Award 1846185, DARPA KAIROS Grant FA8750-19-2-1004, and Munroe \& Rebecca Cobey Fellowship. The views are those of the authors and not the funding agency.

\bibliography{emnlp2020}
\bibliographystyle{acl_natbib}

\appendix

\input{supplement.tex}

\end{document}

%% file: supplement.tex
\section{Appendix}

\begin{table*}[t!]
\small
    \centering
    \begin{tabular}{p{3.7cm}p{3.2cm}lp{5cm}}
    \toprule
         \textbf{Premise} & \textbf{Hypothesis}  & \textbf{Label}  &
         \textbf{Explanation}\\ \midrule
In 870 or 871 he led The Great Summer Army to England. &	In 871 he led The Great Summer Army to England. & neutral & The year can be either 870 or 871, hence we cannot definitely say if it was 871. \\ \midrule

Upon completion, it will rise 64 stories or 711 ft. &	Upon completion, it will rise 711 ft. &	entailment & 64 stories and 711 feet mean the same. "or" is used to establish equivalence between two same things. \\ \midrule

During the shootout, Willis Brooks was killed while a fourth man was seriously wounded. &	During the shootout, Willis Brooks and two others were killed while a fourth man was seriously wounded. & neutral & Whether two others were killed is unknown. \\ \midrule

Gilbert was the freshman football coach of Franklin and Marshall College in 1938. &	Gilbert was the freshman football coach of Franklin College in 1938. &	neutral & Gilbert can be the coach of two colleges with slightly different names ``Franklin and Marshall College" and ``Franklin College".
 \\ \midrule
 
It premiered on 27 June 2016 and airs Mon-Fri 10-11pm IST. &	It premiered on 28 June 2016 and airs Mon-Fri 10-11pm IST. &	contradiction & If it premiered on 27 June, it cannot premier on 28 June. \\ \bottomrule
\end{tabular}
\caption{Some examples from \data{} with gold labels and explanations, used for training the annotators.}
\label{tab:explanation}
\end{table*}

\subsection{Annotation Examples}
In Table \ref{tab:explanation}, we list a subset of examples from \data{}, shown to the annotators with the purpose of training them. In the first example, the ``or'' means ``boolean exclusive-or" while in the second, it is used to establish equivalence between two phrases. The fourth example is a non-boolean usage of ``and'' as it appears as part of a named entity.

\subsection{Automated ``Or'' Adversarial Data Creation}
In this section, we explain the heuristics used to create the ``or'' part of the automated adversarial training data for the IAFT model in Sec.~\ref{adv-data}.
We observe that ``or'' in English is used in sentences in multiple different contexts - (1) establishing exclusivity between options, translating to ``exclusive-or'' in boolean semantics (``He was born in 1970 or 1971.''), (2) establishing equivalence between two words or phrases (``Upon completion, it will rise 64 stories or 711 ft.''),  and (3) ``or'' interpreted as ``boolean or'' (``He can play as a striker or a midfielder.''). Note that the inference label varies between cases (1) and (2) for a particular conjunct operation. For example, in case (1), removal of a conjunct is neutral while for case (2), removal of a conjunct is entailment. Differentiating between these is again challenging due to the lack of any particular trigger in such sentences. We observe that the latter two cases are more frequent in our dataset and thus we heuristically label a pair as entailment when we remove a conjunct and neutral when we add a conjunct. Finally, whenever ``or'' is present in a named entity, we heuristically label the example as neutral.

\begin{table*}[t!]
\small
    \centering
    \begin{tabular}{p{3.6cm}p{3.6cm}lllll}
    \toprule
         \textbf{Premise} & \textbf{Hypothesis} & \textbf{RoBERTa} & \textbf{PA} & \textbf{IAFT} & \textbf{PA-IAFT} & \textbf{Gold} \\ \midrule
         India measures 3214 km from north to south and 2933 km from east to west.
& India measures 3214 km from north to south and 2934 km from east to west. & ent & contra & contra & contra & contra \\ \midrule
He appeared alongside Anthony Hopkins in the 1972 Television series War and Peace. &	He appeared alongside Anthony Hopkins in the 1972 Television series War.	& contra	 & contra & neu &	neu	& neu \\ \midrule
It was released in January 2000 as the lead single from their album "Here and Now". &	It was released in January 2000 as the lead single from their album "Now". &	ent &	ent &	neu	& contra &	contra \\ \midrule
3,000 people died in the days following the earthquakes due to injuries and disease.	& 3,000 people died in the days following the earthquakes due to disease. &	ent &	ent	& ent & ent	& contra \\ \bottomrule
    \end{tabular}
    \vspace{-5pt}
    \caption{Examples from \data{} showing where each model is good at and what is still challenging for all.\vspace{-5pt}}
    \label{tab:error-analysis}
\end{table*}{}

\subsection{BERT-SRL Model}

SRL is the task of predicting the semantic roles for each predicate in the sentence. We follow the standard BIO encoding to denote the span of each argument and model it as a token classification problem. The input to BERT is the tokenized sentence with a $[CLS]$ token at the beginning and a $[SEP]$ token at the end.
\begin{equation}
    [CLS]\hspace{1ex} Sent \hspace{1ex} [SEP] [SEP] \hspace{1ex} Pred \hspace{1ex} [SEP] \nonumber
\end{equation}
\noindent Since BERT converts each word into word pieces, we first propagate the gold SRL tags, which are for each word, to word pieces. Thus, for a word with multiple word pieces, we assign the tag B-ARG to the first word piece, and the tag I-ARG to the subsequent word pieces. For tags of the form I-ARG and O, we assign the original word tag to all the word pieces. The $[CLS]$ and the $[SEP]$ tokens of the input are assigned the tag O. Predicate information is fed to BERT through the segment ids, by assigning 1 to the predicate token and 0 for others. Finally, we add a classification layer at the top and the model is fine-tuned to predict the tag for each token using cross-entropy loss. Unlike \citet{shi2019simple}, our model refrains from using additional LSTM layers at the top, thus making it consistent with other downstream fine-tuning tasks. At inference time, we predict the most probable tag sequence by applying Viterbi Algorithm with two constraints - (1) a sequence cannot start with an I-ARG tag and (2) an I-ARG tag has to be preceded by a B-ARG tag. The tag for each word is taken to be the predicted tag of the first word piece of that word. We train the SRL model on the CoNLL 2005 dataset \cite{Carreras} for 5 epochs with a learning rate of $5*10^-5$ and obtain a near state-of-the-art F1 of $86.23$\% on the dev set.\footnote{The state-of-the-art F1 score for SRL on this dataset is $87.4$\% \cite{ouchi-etal-2018-span}.} The BERT-SRL model's weights are frozen for the NLI task.

\subsection{Experimental Setup}
Our implementation builds on top of the Pytorch implementations \cite{wolf2019transformers} of BERT-base uncased \cite{devlin2018bert} and RoBERTa-base \cite{liu2019roberta}. We train the models for a maximum of 3 epochs using an initial learning rate of $2*10^{-5}$, with linear decay and a weight decay of $0.1$. The dropout probability is chosen to be $0.1$. The size of the predicate-aware representations of the premise and the hypothesis is set to $40$. The maximum sequence length is $128$ for both the NLI models and the SRL model. The random seed used in all the experiments is $42$. Each epoch takes 45 minutes (base models) to 1.5 hours (predicate-aware models) to run on four V100 GPUs. The total number of parameters in our models is similar to that of BERT-base or RoBERTa-base, depending on the choice of the model. All hyperparameters, the amount of adversarial training and the adversarial training algorithm are chosen based on the best average accuracy of \data{} and MNLI dev sets. Batch size and learning rate are manually tuned in the range \{$16$,$32$\} and \{$10^{-5}$, $2*10^{-5}$\} respectively. Following previous works, we report accuracy for all our models.

\subsection{Success and Error Analysis}

In Table \ref{tab:error-analysis}, we present four examples from \data{} and the predictions from each of the models and the target label. In the first example, predicate-aware RoBERTa understands the roles of the predicate ``measures'' and outputs the correct label contradiction, which RoBERTa cannot. The second example shows the effect of adversarial training -- the model learns from adversarial examples that ``War and Peace'' is a named entity and removing a conjunct from it is neutral.\footnote{Note that the gold label is neutral because a person can appear in two Television series with slightly different names.} The third example is however, an exception to the previous rule and only adversarial training fails to make the distinction. On combining predicate-aware RoBERTa with adversarial training, our model learns to associate the predicate ``released'' to its roles and predicts the correct label. Finally, for the fourth example, we find that our model still fails to make the difficult inferences over ``non-boolean and'' cases. Note that there is no obvious trigger like ``total'' in the sentence, thus causing all the models to fail. This also shows that we need a deeper understanding of conjunctive sentential semantics to correctly predict these tricky real-world cases where boolean semantics do not hold.